\pdfoutput=1

	\PassOptionsToPackage{pagebackref=true}{hyperref}
		\documentclass{isprs}
		\usepackage{hyperref}
		\usepackage{subfigure} 

		\usepackage[utf8]{inputenc}
	
		\usepackage{varioref} 
	 
		\usepackage{natbib} 
	
	\usepackage{enumitem}
		\usepackage{graphicx}
		\DeclareGraphicsExtensions{.pdf,.jpg,.png}

		\usepackage[usenames,dvipsnames]{color}
	\usepackage{amsmath}
		\usepackage{algorithm2e}
		\usepackage{microtype}  
		\usepackage{SIunits} 
		\usepackage{listings}
		\usepackage{tabularx} 
		\usepackage{setspace}

		\newcommand{\myimage}[3]
					{
					\begin{figure} [h!]
						\begin{center}
							\includegraphics[width=\linewidth,keepaspectratio]{#1}
							\caption{#2}  
							\label{#3}
							\end{center}
					\end{figure} 
					}

		\newcommand{\myimageHL}[4]
		{
			\begin{figure} [ht!]
				\begin{center}
					\includegraphics[width= #4 \linewidth * 2,keepaspectratio]{#1}
					\caption{#2}  
					\label{#3}
				\end{center}
			\end{figure} 
		}

		\newcommand{\myimageFullPageWidth}[3]
								{
								\begin{figure*}[ht]
									\begin{center}
										\includegraphics[width=\textwidth,keepaspectratio ]{#1}
										\caption{#2}  
										\label{#3} 
										\end{center}
								\end{figure*} 
								}

	\usepackage{rotating}

\title{An octree cells occupancy geometric dimensionality descriptor for massive on-server point cloud visualisation and classification}
\author{Rémi Cura  $^{A}$, Julien Perret $^A$, Nicolas Paparoditis  $^A$}

\address{ $^A$  Université Paris-Est, IGN, SRIG, COGIT \& MATIS, 73 avenue de Paris, 94160 Saint Mandé, France\\
	first\_name.last\_name@ign.fr
	}

\begin{document}
 

 

\abstract{ 	
	Lidar datasets are becoming more and more common. 
	They are appreciated for their precise 3D nature, and have a wide range of applications, such as surface reconstruction, object detection, visualisation, etc.
	For all this applications, having additional semantic information per point has potential of increasing the quality and the efficiency of the application.
	In the last decade the use of Machine Learning and more specifically classification methods have proved to be successful to create this semantic information.
	In this paradigm, the goal is to classify points into a set of given classes (for instance tree, building, ground, other).
	Some of these methods use descriptors (also called feature) of a point to learn and predict its class.
	Designing the descriptors is then the heart of these methods. Descriptors can be based on points geometry and attributes, use contextual information, etc.
	Furthermore, descriptors can be used by humans for easier visual understanding and sometimes filtering.
	In this work we propose a new simple geometric descriptor that gives information about the implicit local dimensionality of the point cloud at various scale.
	For instance a tree seen from afar is more volumetric in nature (3D), yet locally each leaves is rather planar (2D).
	To do so we build an octree centred on the point to consider, and compare the variation of the occupancy of the cells across the levels of the octree.
	We compare this descriptor with the state of the art dimensionality descriptor and show its interest.
	We further test the descriptor for classification within the Point Cloud Server \citep{Cura2016a}, and demonstrate efficiency and correctness results. 
}

\maketitle 

\myimageFullPageWidth{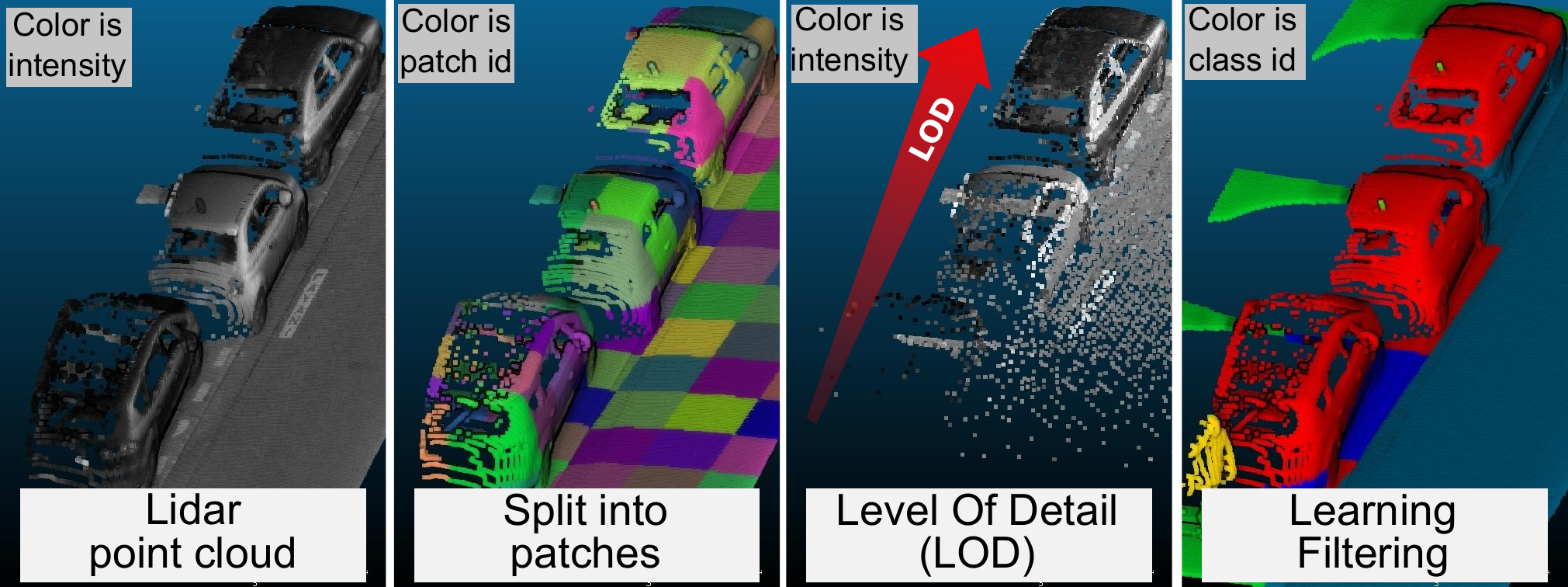}{Graphical Abstract : a Lidar point cloud (1), is split it into patches (2) and stored in a Point Cloud Server, patches are re-ordered to obtain free LOD (3) (a gradient of LOD here), lastly the ordering by-product is a multiscale dimensionality descriptor used as a feature for learning and efficient filtering (4).}{lod.fig:banner_image}


\section{Introduction} 
\subsection{Problem}   
	Democratisation of sensing device have resulted into an expansion of acquired point clouds.
	In the same time, acquisition frequency and precision of the Lidar device are also increasing,
	resulting in an explosion of the number of points.

	Lidar datasets are becoming more and more common. 
	They are appreciated for their precise 3D nature, and have a wide range of applications, such as surface reconstruction, object detection, visualisation, etc.
	
	Semantic information in addition to the raw point data can be very useful for these applications.
	It allows to increase quality and to speed computing.
	For instance a method that reconstruct façade can safely skip the points pertaining to the ground. Similarly, an user visually exploring a dataset may find very useful to isolate points pertaining to trees for instance.
	
	Logically, adding semantic information to point clouds has been researched for a long time by many researchers.
	
	In the last decade the use of Machine Learning and more specifically classification methods have proved to be popular.
	In this paradigm, the goal is to classify points into a set of given classes (for instance tree, building, ground, other).
	Some of this methods uses descriptors (also called feature) for each point that will be leveraged in a training set to learn to associate descriptors with semantic information. 
	Once the association is learned, it can be used to extrapolate semantic classes on similar point clouds. 
	
	The heart of such approaches are then to design appropriate descriptors that will enable to accuratly and efficiently disambiguate between classes.
	Many different descriptors have been used, based on geometry or other attributes of points, using or not the context of the point, etc.
	We refer to the very recent thesis of \cite{Weinmann2016}.
	
	Recently deep learning methods potentially allow to learn features on the fly (see \cite{Huang2016}), bypassing the need to craft descriptors, but also introducing new trade-off in the process (necessary training set and prior knowledge of similar point clouds).
	
	Of course machining learning approach requires substantial training sets, and may fail if the processed point cloud is too different from the learned point clouds.
	Furthermore, machine learning methods are sophisticated and require significant computing time.
	However deep learning methods have been applied to information extracted from point clouds, such a 2D images(\cite{Boulch2017}), voxels (\cite{Tchapmi2017}), and graph of relation between superpoints (\cite{Landrieu2017}).
	
	In this work we focus on a new simple geometric descriptor that gives information about the implicit local dimensionality of the point cloud, regardless of prior knowledge, methods of acquisition and scale.
	This descriptor is based on the octree cells occupancy of a point cloud.
    It is designed with massive point clouds and scaling in mind,
	which means that it is included in a global Level Of Detail approach,
	and is computed on groups of points rather than on individuals points.
	
	Our aim is to provide a basic descriptor that has a direct geometrical interpretation, is extremely fast to compute and scales well due to its hierarchical nature.
	This descriptor can then be used for visualisation or as a preprocessing step for classification and reconstruction. 
	
	We compare this descriptor with the state of the art. 
	We demonstrate the potential of th descriptor to perform efficient patch classification within the Point Cloud Server \citep{Cura2016a}.

\subsection{Contribution}

	All the methods are tested on billions scale point cloud, and are Open Source for sake of reproducibility test and improvements.
	
	Our main contribution is an efficient and robust geometric dimensionality descriptor with a scale parameter that is local rather than global.
	Our second contribution is to explore the interest of classification of groups of points (patches) rather than points for massive point clouds, and estimate the trade-off associated to this approach. 
	 
\subsection{Plan}
	This work follows a classical plan of Introduction Method Result Discussion Conclusion (IMRAD).
	Section~\ref{dim.sec:method} presents the geometric dimensionality descriptor, and how this can leveraged for classification.  
	Section~\ref{dim.sec:result} reports on the experiments validating the descriptor interest and how it compares to the state of the art geometric dimension descriptors. 
	Finally, the details, limitations, and potential applications are discussed in Section~\ref{dim.sec:discussion}.



\section{Method}
	\label{dim.sec:method}
	
	In this section, we first present the Point Cloud Server (section \ref{dim.method.PCS})(PCS \cite{Cura2016})
	that this article extends. Then we introduce a dimensionality descriptor that is based on octree cells occupancy (\ref{dim.method.dimdescriptor}). 
	This descriptor can be used in the PCS to perform density correction and classification at the patch level (\ref{dim.method.classif}).
	This classification can be directly transferred to points, or indirectly exploited in a pre-filtering step.
	
	\subsection{The Point Cloud Server}
	\label{dim.method.PCS}
		\myimage{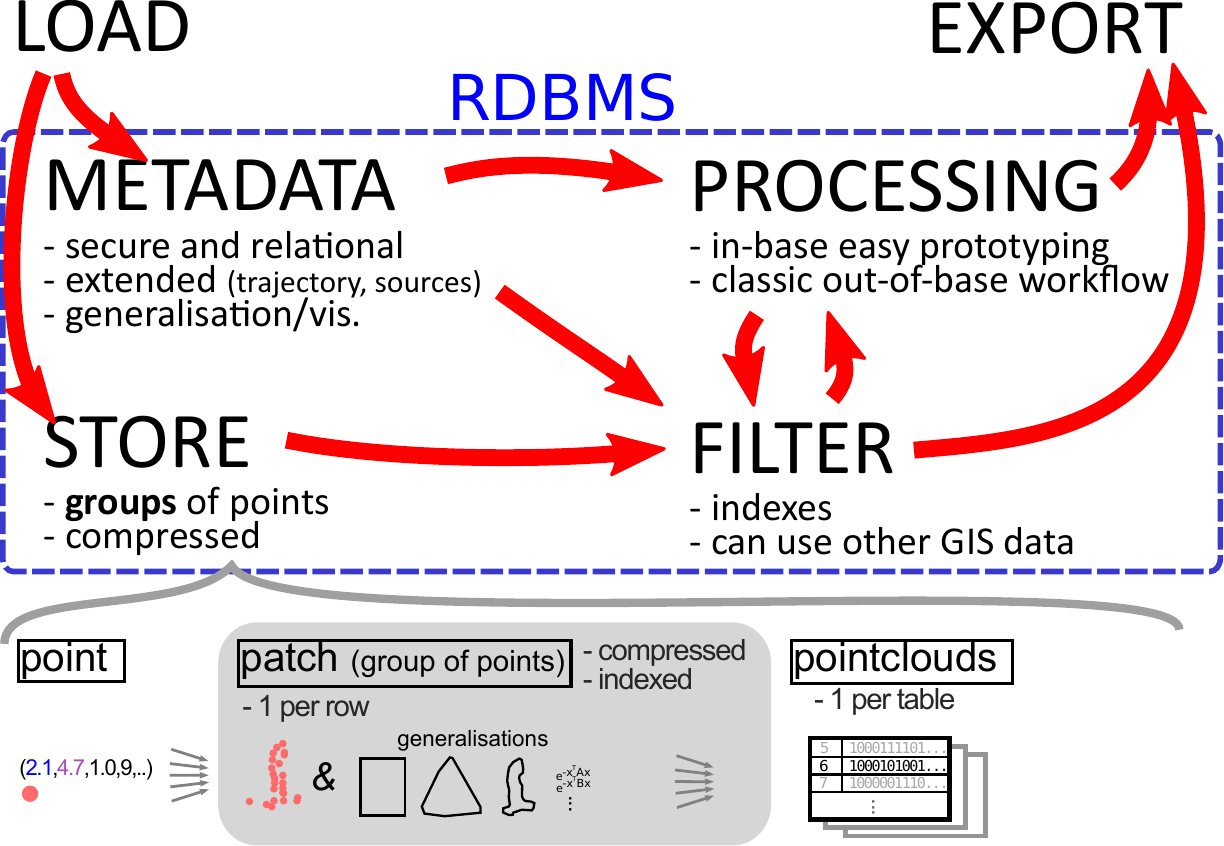}{Overall and storage organisations of the Point Cloud Server.}{dim.fig:PCS}
		
		Our method strongly depends on using the Point Cloud Server described in \cite{Cura2016},
		therefore we introduce its principle and key relevant features (see figure \ref{dim.fig:PCS}).
		
		The PCS is a complete and efficient point cloud management system based on a database server that works on groups of points rather than individual points.
		This system is specifically designed to solve all basics needs of point cloud users:
		fast loading, compressed storage, powerful filtering, easy data access and exporting, and integrated processing.
		
		The core of the PCS is to store groups of points (called patches) that are multi-indexed (spatially, on attributes, etc.), and represented with different generalisation depending on the applications.
		Points can be grouped with any rules.
		In this work, the points are regrouped spatially by cubes $1 \metre$ (Paris) or $50 \metre$ (Vosges) wide.
		
		All the methods described in this work are applied on patches.

	\subsection{ A local multi-scale dimensionality descriptor: the octree cells occupancy descriptor} 
		\label{dim.method.dimdescriptor}
		\subsubsection{Principle}
		The goal is to estimate the geometric dimensionality of a group of points (is the group of points more 1D, 2D, 3D in nature?). For instance a door would be mostly 2D, and a telephone line mostly 1D.
		Our approach is multi-Level Of Details (LOD), because we consider that the geometric dimensionality of a group of points depends on the scale/the level of detail.
		
		Lets consider a group of points, we build an octree where the cell of the first level is a cube containing all the points. In the Point Cloud Server, the cube size is chosen for storage efficiency (between 50m and 1m of side length) and aligned on a 3D grid.
		For a given level $L_i$ level of the octree, we note the number of cells that are occupied by one or more points.
		Doing so for $n$ levels create a feature vector $ppl=(O_1,O_2,..O_n)$ where $O_i$ is the number of cells of the level $i$ of the octree that contains 1 or more points.

		For instance, given a level $L_i$, a centred line would occupy $2^{L_i}$ cells, a centred plan $4^{L_i}$ cells, and a volume all of the cells ($8^{L_i}$ cells).
		Thus simply by counting the number of occupied cells we get an idea of the geometric dimensionality of the group of points for a given level (See Figure \ref{dim.fig:dim_descriptor}). 
		\myimage{"./illustrations/chap2/dim_descriptor/dim_descriptor"}{Octree cells occupancy is a basic dimensionality descriptor: a 3D line, 2D surface or volume occupy a different amount of cells.}{dim.fig:dim_descriptor}.
		This occupancy is only correctly estimated when the patch is fully filled and dimensionality homogeneous. 
		However, we can also characterize the dimensionality $Dim_{LODDiff}$ by the way the occupancy evolves (difference mode).
		Indeed, a line occupying $k$ cells of an octree at level $L_i$ will occupy $2*k$ cells at the level $L_{i+1}$, if enough points (see Fig. \ref{dim.fig:analysing_tree} ).
		
		\myimage{"./illustrations/chap2/comparing_dim_desc/analysing_tree/analysing_tree"}{Illustration of the octree cell occupancy evolution for a point cloud acquisition of a real life tree (Dimension is embedded it the power of 2). Depending on the Level (thus the scale), the geometric dimensionality of the object goes from 3D to 1D.}{dim.fig:analysing_tree}

		The Figure \ref{dim.fig:lod-common-objects} illustrate this. Typical parts of a street in the Paris dataset were segmented: a car, a wall with window, a 2 wheelers, a public light, a tree, a person, poles and piece of ground including curbs.
		\\
		Due to the geometry of acquisition and sampling, the public light is almost a 3D line, resulting in the occupation of very few octree cells.
		A typical number of octree cells per level for a public light patch would then be $(1,2,4,8,...)$, which looks like a $(2^1)^L$ function.
		A piece of ground is often extremely flat and very similar to a planar surface,
		which means that the occupied octree cells could be a $(1,4,16,64...)$, a $(2^2)^L$ function.
		Lastly a piece of tree foliage is going to be very volumetric in nature,
		due to the fact that a leaf is about the same size as point spacing and is partly transparent to laser (leading to several echo).
		Then a foliage patch would typically be $(1,8,64...)$ (if enough points), so a $(2^3)^L$ function.
		(Tree patches are in fact a special case, see \ref{dim.result.dim_failure}).
		
		\myimageHL{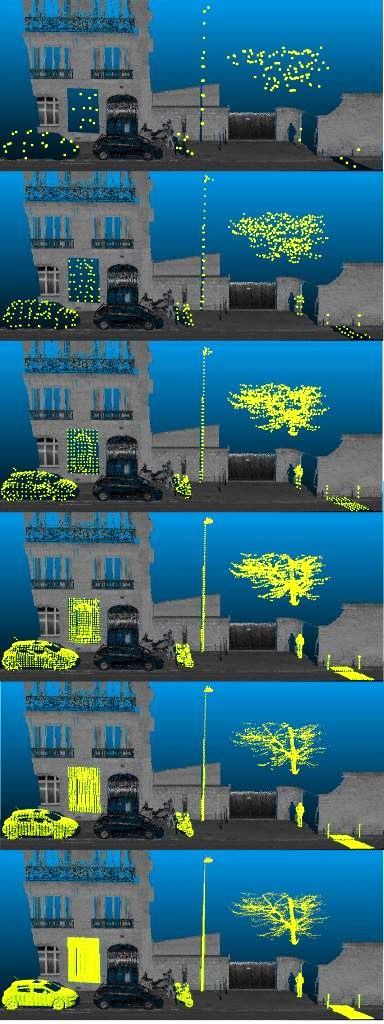}{All successive levels of LOD from \cite{Cura2016} for common objects (car, window, 2 wheelers, light, tree, people, pole, ground), color is intensity for other points.}{dim.fig:lod-common-objects}{0.95}
		
		\subsubsection{Link to the MidOc ordering}
		In \cite{Cura2016} we introduced a new method for Level Of Detail of massive point cloud. 
		The main idea was to store implicitly the LOD within the order of points. 
		Thus, the method relies on ordering the points with the most important (geometrically speaking first). This order method called MidOc is also build around an Octree.	
		When using the Implicit LOD MidOc building process, the number of chosen points per level can be stored.
		Each ordered patch is then associated with a vector of number of points per level $ppl=(N_{L_{1}},..,N_{L_{\text{max}}})$.
		The number of picked point for $L_i$ is almost the voxel occupancy for the level $i$ of the corresponding octree.
		Almost, because in MidOc points picked at a level do not count for the next Levels.
		Occupancy over a voxel grid has already been used as a descriptor (See \cite{Bustos2005}).
		However we can go a step further.
		For the following we consider that patches contain enough points and levels are low enough so that the removing of picked points has no influence. 
		Thus, by comparing $ppl[L_i]$ to theoretical $2^{L_i}, 4^{L_i}, 8^{L_i}$ we retrieve a dimensionality feature  $Dim_{LOD}[i]$  about the dimensionality of the patch at the level $L$ (See Figure \ref{dim.fig:dim_descriptor}).  
		This occupancy is only correctly estimated when the patch is fully filled and homogeneous. 
		However, we can also characterize the dimensionality $Dim_{LODDiff}$ by the way the occupancy evolves (difference mode).
		Indeed, a line occupying $k$ cells of an octree at level $L_i$ will occupy $2*k$ cells at the level $L_{i+1}$, if enough points.

		\subsubsection{Comparing octree cells occupancy descriptor with covariance - based descriptors}
		
		A sophisticated per-point dimensionality descriptor is introduced in \cite{Demantke2014, Weinmann2015}, then used to find optimal neighbourhood size.
		This approach is very different in the approach.
		First that this feature is computed for each point (thus is extremely costly to compute).
		On the opposite, in the PCS we compute feature at the patch level, we do not need to find the scale at which compute dimensionality, the descriptor is computed on the whole patch.

		This dimensionality descriptors ($Dim_{cov}$) relies on computing covariance of points centred to the barycentre (3D structure tensor), then a normalisation of covariance eigen values.
		As such, the method is similar, and has the same advantages and limitation, as the Principal Component Analysis (See \cite{Shlens2014} for a reader friendly introduction).
		It can be seen as fitting a 3D ellipsoid to the points.
		
		First this method is sensible to density variations because all the points are considered for the fitting. 
		As opposite to our hypothesis,
		this method considers implicitly that density holds information about the nature of sensed objects. 
		Second, this methods only fits one ellipse, which is insufficient to capture complex geometric forms. 
		Last, this method is very local and does not allow to explore different scale for a point cloud as a whole. Indeed this method is classically used on points within a growing sphere to extend the scale.
		
		The second difference is more methodological, as typically the scale of the feature would be the radius into which consider points to compute feature. Thus it is not possible to analyse a large group of points at a geometrically fine scale.
		In the opposite, in the case of the octree cells occupancy, the scale is actually directly related to the geometric size of the octree cells.
		
			\myimageHL{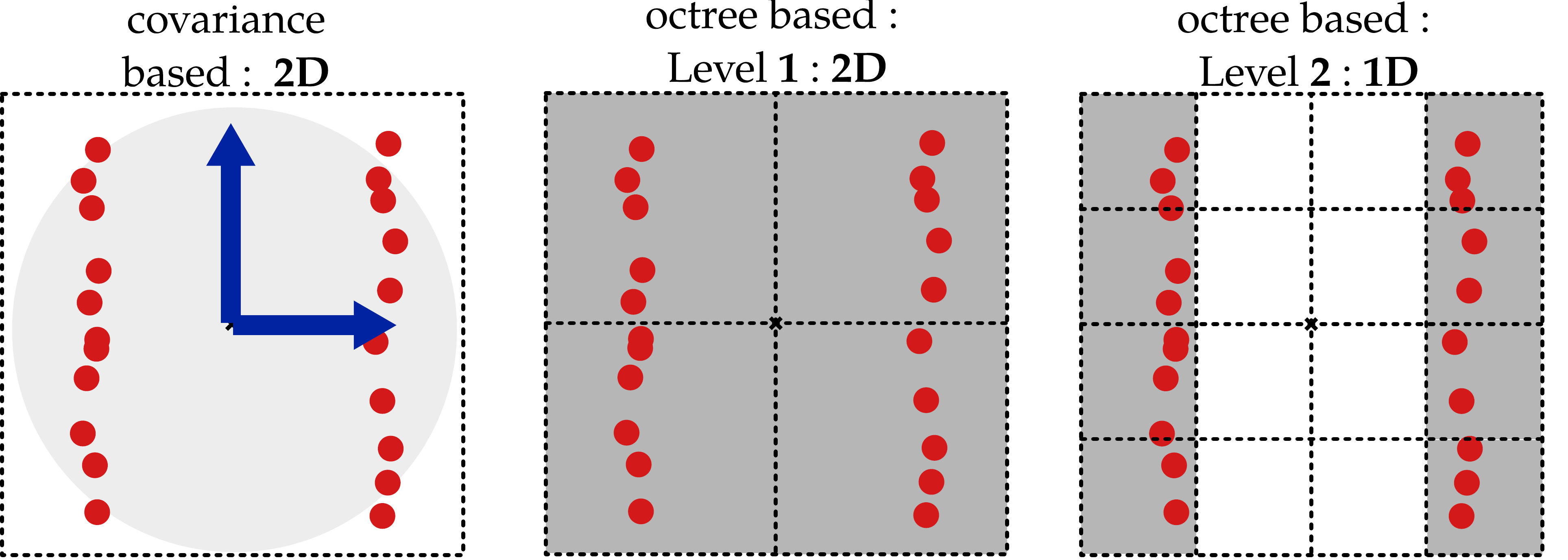}{Covariance-based geometric dimensionality is estimated similarly to a PCA, the scale is the size of the neighbourhood radius. In the opposite octree cells occupancy geometric dimensionality descriptor scale is in fact the scale to which the points are analysed.}{dim.fig:covariance_vs_octree}{0.95}
		
		We compute both dimensionality descriptor and then compare them for the Paris dataset.
		
		\subsubsection{Octree cells occupancy descriptor as a feature for classification}

		Because $x_1 \rightarrow (2^1)^x$,
		$x_2 \rightarrow (2^2)^x$, $x_3 \rightarrow (2^3)^x$ diverge very fast,
		we only need to use few levels to have a quite differentiating descriptor.
		
		For instance, using $L=2$, we have $x_i=[4,16,64]$ , which are very distinguishable values, and don't require a total density above $70$ points per patch.  
		As long as the patch contains a minimal number of points, the descriptors is density and scale invariant.
		Lastly a mixed result (following neither of the $x_i \rightarrow (2^i)^x$ function) can be used as an indicator that the patch contains mixed geometry, either due to nature of the objects in the patch, or due to the way the patch is defined (sampling).
		
		Although it might be possible to go a step further and decompose a patch $ppl$ vector on the base of $x_i \rightarrow (2^i)^x, i \in [1..3]$, the direct and exact decomposition can't be used because the decomposition might depends on $L_i$. For instance a plane with a small line could appear as a plan for $L_1$ and $L_2$, and starts to appear differently over $L_3$ and higher level. In this case, an Expectation-Maximization scheme might be able to decompose robustly too separate the points into more dimensionally coherent groups.
   
	\subsection{Rough patch classification with the Point Cloud Server}
		\label{dim.method.classif}
		\subsubsection{Principle}
		
		We propose to perform patch classification using the Point Cloud Server and the previously introduced octree cells occupancy, along with other basic descriptors, using a Random Forest classifier.
		Following the position of the PCS towards abstraction, the classification is performed at the patch level and not at the point level. 
		This induces a massive scaling and speeding effect, at the cost of introducing quantization error.
		Indeed, compared to a point classification, a patch may contain points belonging to several classes (due to generalisation), yet it will only be classified in one class, thus the "quantization" error.
		
		Because patch classification is so fast and scales so well,
		the end goal can be however slightly different than for usual point classification.

		Patch classification can be used as a fast pre-process to another slower point classification,
		be it to speed it (an artificial recall increase for patch classification may be needed, see Figure \ref{dim.fig:recall-increase}), or to better a point classification.
		The patch classification can provide a rough classification.
		Based on that the most adapted point classifier is chosen 
		(similarly to Cascaded classifiers),
		thus improving the result of the final point classification.
		For instance a patch classified as urban object would lead to chose a classifier specialized in urban object, and not the general classifier.
		This is especially precious for classes that are statistically under-represented (i.e. pedestrian, urban furniture).
		 
		Patch classification may also be used as a filtering pre-process for applications that only require one class. 
		Many applications only need one class, and do not require all the points in it, but only a subset with good confidence.
		For this it is possible to artificially boost the precision (by accepting only high confidence prediction).
		For instance computing a Digital Terrain Model (DTM) only requires ground points.
		Moreover, the ground will have many parts missing due to objects,
		so using only a part of all the points will suffice anyway. 
		The patch classifier allow to find most of the ground patch extremely fast.
		Another example is registration.
		A registration process typically require reliable points to perform mapping and registration.
		In this case there is no need to use all points,
		and the patch classification can provide patches from ground and façade with high accuracy
		(for point cloud to point cloud or point cloud to 3D model registration),
		or patches of objects and trees (for points cloud to landmark registration).
		In other applications, finding only a part of the points may be sufficient, for instance when computing a building map from façade patches.

		Random Forest method started with \cite{Amit97shapequantization}, theorized by \cite{Breiman2001} and has been very popular since then. They are for instance used by \cite{Golovinskiy2009} who perform object detection, segmentation and classification. They analyse separately each task on an urban data set, thus providing valuable comParison. Their method is uniquely dedicated to this task, like \cite{Serna2014} who provide another method and a state of the art of the segmentation/classification subject.
		Both of this methods are in fact 2D methods, working on an elevation image obtained by projecting the point cloud. However we observe that street point clouds are dominated by vertical surfaces, like building (about 70\% in Paris data set). Our method is fully 3D and can then easily be used to detect vertical object details, like windows or doors on buildings.

		\subsubsection{Rough patch classification details}  
		\paragraph{Features}
		The first descriptor is $ppl$, the octree cells occupancy dimensionality descriptor, usually produced by the MidOc ordering (see Section \ref{dim.method.dimdescriptor}), or computed independently using fast octree building by points ordering.
		We use the number of points for the level $[1..4]$. For each level $L$, the number of points is normalized by the maximum number of points possible ($8^L$), so that every feature is in $[0,1]$.
		
		\label{dim.method.classification.other_feature}
		We also use other simple features that require very limited computing (avoiding complex features like contextual features). 
		Due to the PCS patch compression mechanism,
		min, max, and average of any attributes of the points are directly available.
		Using the LOD allows to quickly compute other simple feature, like the 2D area of points of a patch (points being considered with a given diameter). 
		
		\paragraph{Dealing with data set particularities}
		The Paris data set classes are organized in a hierarchy (100 classes in theory, 22 populated).
		The rough patch classifier is not designed to deal with so many classes,
		and so a way to determine what level of hierarchy will be used is needed.
		We propose to perform this choice with the help of a graph of similarity between classes (See Fig. \ref{dim.fig:ppl-separator-power} and \ref{dim.fig:class-clustering-all-features}

		\label{dim.method.classification.spectral_layout}
		We first determinate how similar the classes are for the simple dimensionality descriptors, classifying with all the classes, and computing a class to class confusion matrix.
		This confusion matrix can be interpreted as an affinity between class matrix, and thus as a graph.
		We draw the graph using a spectral layout (\cite{Networkx2014}),
		 which amounts to draw the graph following the first two eigen vector of the matrix (Similar to Principal Component Analysis).
		Those two vectors maximize the variance of the data (while being orthogonal), and thus best explain the data.
		This graph visually helps to choose the appropriate number of classes to use.
		A fully automatic method may be used via unsupervised clustering approach on the matrix 
		(like The Affinity Propagation of \cite{Frey2007}).
		
		Even when reducing the number of classes, the Paris dataset if unbalanced (some class have far less observations than some others).
		We tried two classical strategies to balance the data set regarding the number of observation per class.
		The first is under-sampling big classes : we randomly under-sample the observations to get roughly the same number of observation in every class.
		
		The second strategy is to compute a statistical weight for every observation based on the class prevalence. 
		This weight is then used in the learning process when building the Random Forest.
		
		\subsubsection{Using the confidence result from the classifier} 
		\label{dim.method.classification.using_confidence}
		Contrary to classical classification method, we are not exclusively interested in precision and recall per class, but also by the evolution of precision when prediction confidence varies.
		
		In fact, for a filtering application, we can leverage the confidence information provided by the Random Forest method to artificially boost precision (at the cost of recall diminution). We can do this by limiting the minimal confidence allowed for every prediction.
		Orthogonally, it is possible for some classes to increase recall at the cost of precision by using the result of a first patch classification and then incorporate in the result the other neighbour patches. 
		
		We stress that if the goal is to detect objects (and not classify each point), this strategy can be extremely efficient.
		For instance if we are looking for objects that are big enough to be in several patches (e.g. a car).
		In this case we can perform the classification (which is very fast and efficient), then keep only highly confident predictions, and then use the position of predictions to perform a local search for car limits.
		The classical alternative solution would be to perform a per point classification on each point, which would be extremely slow.
		 
	

 \section{ Result }
	 \label{dim.sec:result}
 	\subsection{Introduction to results}
 		We design and execute several experiments in order to test the descriptors for a random forest classifier on two large data sets, proving their usefulness.
 		We analyse the potential of this descriptors, and what it brings when used in conjunction to other simple descriptors.
 		
 		The base DBMS is \cite{PostgreSQL2014}. The spatial layer \cite{PostGIS2014} is added to benefits from generic geometric types and multidimensional indexes. The specific point cloud storage and function come from \cite{pgPointCloud2014}. 
 		The MidOc is either plpgsql or made in python with \cite{SciPy2014}. 
 		The classification is done with \cite{scikit-image}, and the network clustering with \cite{Networkx2014}.
 		Timings are only orders of magnitude due to the influence of database caching.
 	
		\myimageHL{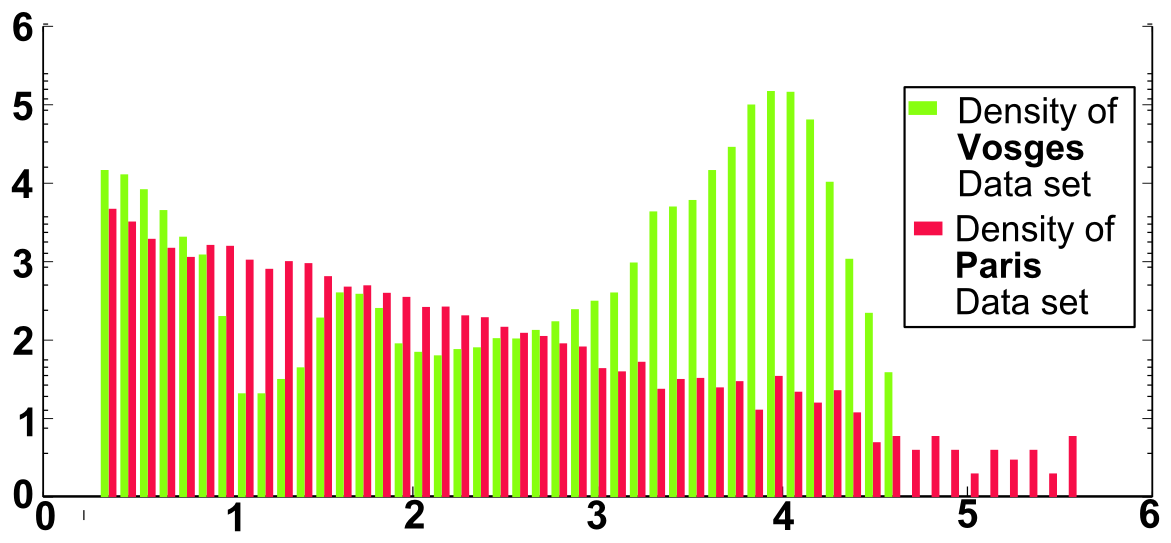}{ Histogram of number of points per patch, with a logarithmic scale for X and Y axis}{dim.fig:hist-density-dataset}{0.95}
	 
 		We use two data sets. There were chosen as different as possible to further evaluate how proposed methods can generalise on different data (See Figure fig:hist-density-dataset for histogram of patch density ). 
 		The first data set is \cite{IQmulus2014} (Paris data set), an open source urban data set with varying density, singularities, and very challenging point cloud geometry. 
 		Every point is labelled with a hierarchy of 100 classes.
 		The training set is only 12 millions points.
 		Only 22 classes are represented. We group points in $1 \cubic \metre$ cubes.
 		The histogram of density seems to follow an exponential law (See figure \ref{dim.fig:hist-density-dataset}), the effect being that many patches with few points exist. 
 		
 		We also use the Vosges data set, which is a very wide spread, aerial Lidar, 5.5 Billions point cloud. 
 		Density is much more constant at 10k pts/patch .
 		A vector ground truth about surface occupation nature (type of forest) is produced by the French Forest Agency. Again the classes are hierarchical, with 28 classes.
 		We group points in 50 $\times 50 \meter$ squares.

	\subsection{Using the Point Cloud Server for experiments}
		All the experiments are performed using a Point Cloud Server (cf \cite{Cura2014}).
		The key idea are that point clouds are stored inside a DBMS (PostgreSQL), as patch. Patch are compressed groups of points along with some basic statistics about points in the group.
		We hypothesize that in typical point cloud processing work-flow, a point is never needed alone, but almost always with its surrounding points.
	
		Each patch of points is then indexed in an R tree for most interesting attributes (obviously X,Y,Z but also time of acquisition, meta data, number of points, distance to source, etc.)
			
		Having such a meta-type with powerful indexes allows use to find points based on various criteria extremely fast. (order of magnitude : ms). 
		As an example, for a 2 Billion points dataset, we can find all patches in few milliseconds having : 
		 - between -1 and 3 meters high in reference to vehicle wheels
		 - in a given 2D area defined by any polygon 
		 - acquired between 8h and 8h10 - etc.
		 
		The PCS offers an easy mean to perform data-partition based parallelism. We extensively use it in our experiments.

	\subsection{Multi-scale local Dimensionality descriptor}
		\label{dim.result.dim_descriptor}
		We test the dimensionality descriptor ($ppl$) in two ways.
		First we compare the extracted ($Dim_{LOD}$) to the classical structure tensor based descriptor ($Dim_{cov}$).
		Second we assess how useful it is for classification, 
		by analysing how well it separates classes, and how much it is used when 
		several other features are available.
		
		\subsubsection{Comparing LOD-based descriptor with Structure tensor-based descriptor}
		We compute $Dim_{cov}$ following the indications of \cite{Weinmann2015} to get $p_{dim}->[0..1]^3$, i.e. the probability to belong to [1D,2D,3D].
		We convert this to $Dim_{cov}$ with $Dim_{cov}=\sum_{i=1}^{3}{i*p_{dim}[i]}$.
		
		Optionally, we test a filtering option so that the maximum distance in biggest two dimensions is more equivalent. However this approach fails to significantly improve results.
		
		We test several method to extract $Dim_{LOD}$ from $ppl$.
		The first method is to compute $ Dim_{LODs}[i] = log2(ppl[i])/i$,
		which gives the simple geometric dimension for each level.
		The second method is the same but work on occupancy evolution, with
		$Dim_{LODd}[i] = log2(ppl[i]/ppl[i-1])$ (discarding $L_0$).
		In both case the result is a geometric dimension between 0 and 3 for each Level.
		We use both indices to fusion the dimensionality across Levels (  working on $Dim_{LODA} = Dim_{LODs}\bigcup Dim_{LODd}$).
		The first method uses a RANSAC (\cite{SciPy2014} implementation of \cite{Choi2009}) 
		to find the best linear regression. The slope gives an idea of confidence (ideally, should be 0),
		and the value of the line at the middle of abscissa is an estimate of $Dim_{LOD}$.
		The second method robustly filters $Dim_{LODA}$ based on median distance to median value and
		average the inlier to estimate $Dim_{LOD}$.
		
		$Dim_{cov}$ and $Dim_{LOD}$ are computed with in-base and out-of-base processing, the latter 
		being executed in parallel (8 workers).
		For 10k patches, 12 \mega pts, retrieving data and writing result accounts for $48\second$, computing $Dim_{LOD}$ to $8\second$, $Dim_{cov}$ to $64\second$. Computing $ppl$ (which is multi-scale) using a linear octree takes between $58$ $L_6$ and $85 \second$ $L_8$.
		 (
		\myimageFullPageWidth{"./illustrations/chap2/comparing_dim_desc/success_case"}{$Dim_{LOD}$ and $Dim_{cov}$ are mostly comparable, except for few patches (5\%, coloured)}{dim.fig:success_case}{0.75}

		Comparing $Dim_{cov}$ and $Dim_{LOD}$ is not straightforward because the implicit definition of dimension is very different in the two methods.
		We analyse the patches where $\lvert Dim_{LOD} -  Dim_{cov}\rvert <=0.5$. 
		0.5 is an arbitrary threshold,
		but we feel that it represents the point above which descriptors will predict unreconcilable dimensions.
		Those patches represent 93\% of the data set (0.96 \% of points), with a correlation of 0.80.
		Overall the proposed dimensions are similar for the majority of patch, especially for well filled 1D and 2D patches (See Fig. \ref{dim.fig:success_case}).

		\myimageHL{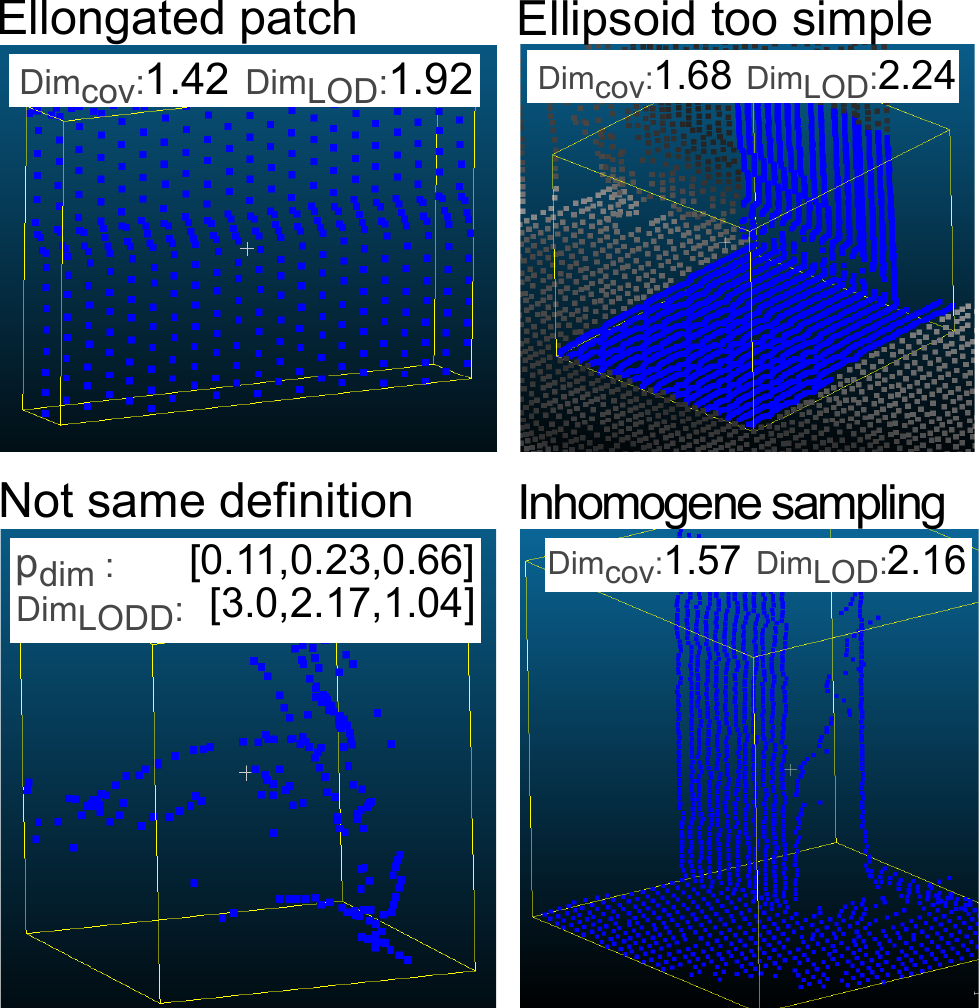}{Representative patches for $\lvert Dim_{LOD}$-$Dim_{cov}>0.5\rvert$. Most differences are explained by $Dim_{cov}$ limitations (See \ref{dim.result.dim_failure}).}{dim.fig:explaining_failure_case}{0.95}
		
		We analyse the 684 remaining patches to look for possible explanations of the difference in dimension (See Fig. \ref{dim.fig:explaining_failure_case}).
		
		We consider the following four main sources of limitations from $Dim_{cov}$.
		\begin{itemize}[noitemsep,topsep=0pt,parsep=0pt,partopsep=0pt]
			\label{dim.result.dim_failure}
			\item Elongated patch.\\
				$Dim_{cov}$=1.42
				,$Dim_{LOD}$=1.92. 
				If the patch is not roughly a square,$Dim_{cov}$ gives a bad estimation as it is biased by the un-symmetry of point distribution.
			\item Ellipsoid too simple.\\
				$Dim_{cov}$=1.68,
				$Dim_{LOD}$=2.24.
				$Dim_{cov}$ fits an ellipsoid, which can not cope with complex objects, especially when the barycentre does not happen to be at a favourable place. 
			\item Coping with heterogeneous sampling.\\
				$p_{dim}$=[0.56,0.32,0.12],
				$Dim_{cov}$=1.57,
				$Dim_{LOD}$=2.16.\\
				$Dim_{cov}$ is sensitive to difference in point density. The points on the bottom plan are much 3 times less dense than in the vertical plan, leading to a wrong estimate.
			\item Definition of dimension different.\\
				General:$Dim_{cov}\in[1.2,2.6],Dim_{LOD}\in[1.7..2.7]$\\
				This patch: $p_{dim}$=[0.11, 0.23, 0.66]\\
				$ppl$=[1,8,36,74..],
				$Dim_{LODD}$=[3.0,2.17,1.04].
				Trees are a good example of how the two descriptors rely on a different dimension definition. For $Dim_{cov}$ points may be well spread out, so usually $p_{3D}$ is high.
				Yet, tree patches are also subject to density variation, and may also be elongate, which renders $Dim_{cov}$ very variable.
				On the opposite, $Dim_{LOD}$ considers the dimensionality at different scale (See Fig. \ref{dim.fig:analysing_tree}). From afar a tree-patch is volumetric, at lower scal, it seems planar (leaf and small sticks form rough plans together). Lastly at small scale, the tree looks linear (sticks). 
		\end{itemize}

		\subsubsection{Usefulness of rough descriptor for classification}
		\myimageHL{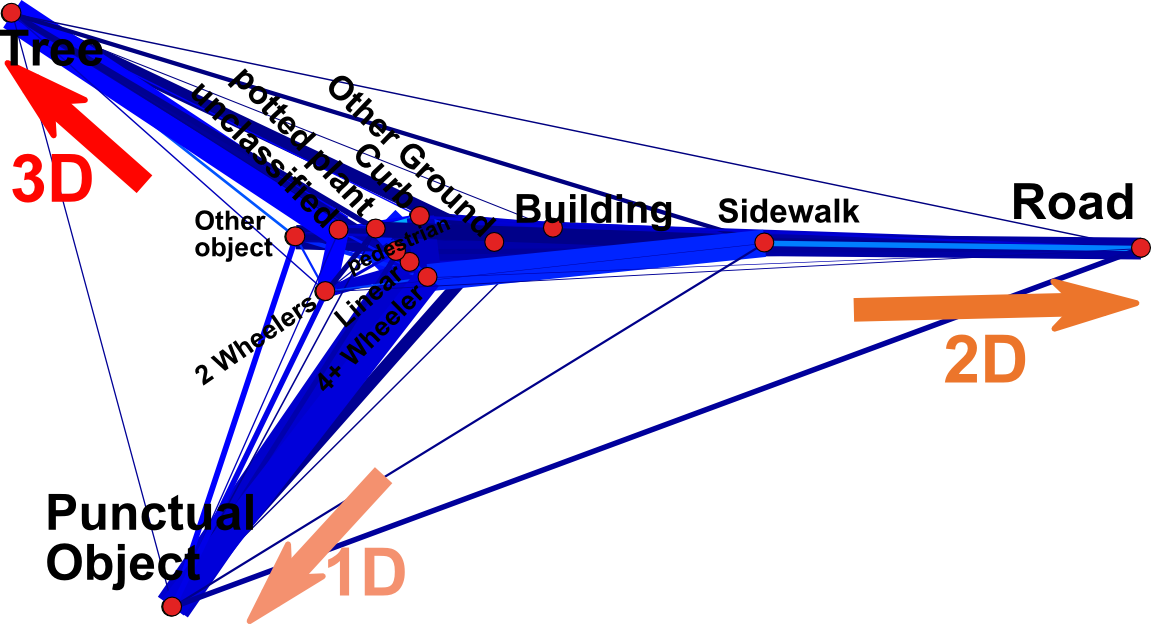}{Spectral clustering of confusion matrix of Paris data set classification using only $ppl$ descriptor. Edge width and colour are proportional to affinity. Node position is determined fully automatically. Red-ish arrows are manually placed as visual clues.}{dim.fig:ppl-separator-power}{0.95}
		
		Using only the $ppl$ descriptor, a classification is performed on Paris data set, then a confusion matrix is computed.
		We use the spectral layout (see Section \ref{dim.method.classification.spectral_layout}) to automatically produce the graph in Figure \ref{dim.fig:ppl-separator-power}. 
		We manually ad 1D,2D and 3D arrows.
		On this graph, classes that are most similar according to the classification with $ppl$ are close. The graph clearly present an organisation following 3 axis. Those axis coincide with the dimensionality of the classes. For instance the "tree" class as a strong 3D dimensionality. The "Punctual object" class, defined by "Objects which representation on a map should be a point", is strongly 1D (lines), with object like bollard and public light. The "Road" class is strongly 2D, the road being locally roughly a plan. The centre of the graph is occupied by classes that may have mixed dimensionality, for instance "4+ wheeler" (i.e. car) may be a plan or more volumetric, because of the $1 \cubic \metre$ sampling.
		"Building" and "sidewalk" are not as clearly 2D as  the "road" class. Indeed, the patch of "sidewalk" class are strongly mixed (containing 22\% of non-sidewalk points, See figure \ref{dim.fig:result-Paris}). The building class is also not pure 2D because building façade in Paris contains balcony, building decoration and floors, which introduce lots of object-like patches, which explain that building is closer to the object cluster. (See Figure \ref{dim.fig:lod-common-objects} for instance).
		The dimensionality descriptor clearly separates classes following their dimensionality, but can't separate classes with mixed dimensionality.

		To further evaluate the dimensionality descriptor, we introduce other classification features (see \ref{dim.result.classification}), perform classification and compute each feature importance.
		The overall precision and recall result of these classification is correct, and the $ppl$ descriptor is of significant use (See Figure \ref{dim.fig:result-Paris} and \ref{dim.fig:result-Vosges}), especially in the Vosges data set. The $ppl$ descriptor is less used in Paris data set, maybe because lots of classes can not really be defined geometrically, but more with the context.
		  
	\subsection{Patch Classification}
		\label{dim.result.classification}
		 
		\subsubsection{Introducing other features}
		The dimensionality descriptor alone cannot be used to perform sophisticated classification,
		because many semantically different objects have similar dimension 
		(for instance, a piece of wall and of ground are dimensionally very similar
		, yet semantically very different).
		We introduce additional simple features for classification (See Section \ref{dim.method.classification.other_feature}). All use already stored patch statistics, and thus are extremely fast to compute.
		(P : for Paris , V : for Vosges: 
		- average of altitude regarding sensing device origin(P)
		- area of $patch\_bounding\_box$ (P) : 
		- patch height (P)
		- $points\_per\_level$ ($ppl$), level 1 to 4 (P+V)
		- average of intensity (P+V)
		- average of $number\_of\_echo$ (P+V) 
		- average Z (V)
		
		For Vosges data set, we reach a speed of 1 \mega points\per \second \per worker to extract those features.
		\subsubsection{Classification Setting} 
		Undersampling and weighting are used on the Paris dataset. First undersampling to reduce the over dominant building class to a 100 factor of the smallest class support. Then weighting is used to compensate for differences in support. 
		For the Vosges data set only the weighting strategy is used. 
		The weighting approach is favoured over undersampling because it lessen variability of results when classes are very heterogeneous.
		
		To ensure significant results we follow a K-fold cross-validation method. We again compute a confusion matrix (i.e. affinity between classes) on the Paris data set to choose which level of class hierarchy should be used.
		fig:class-clustering-all-features

		\subsubsection{Analysing class hierarchy} 
		\myimageHL{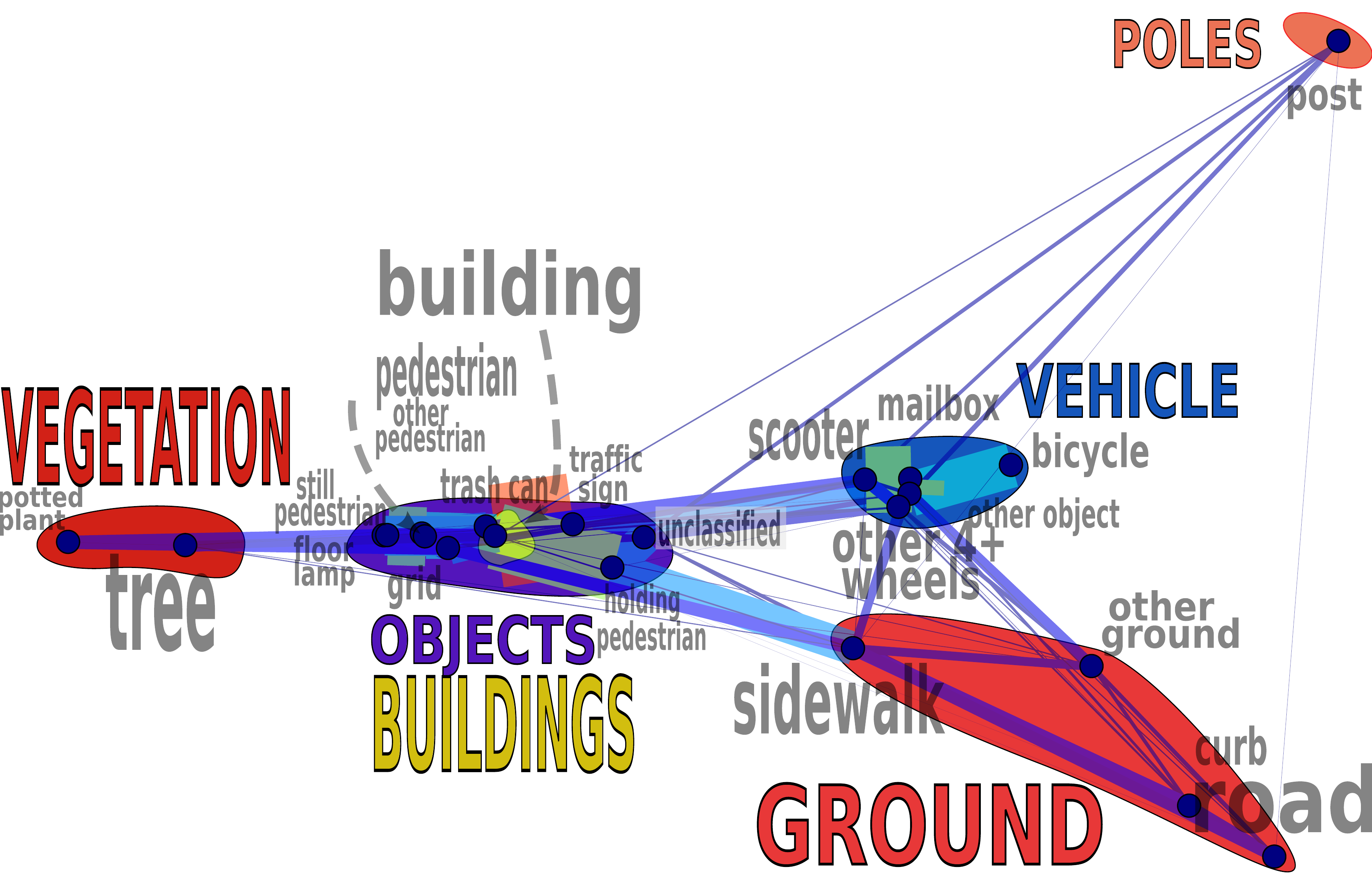}{Result of automatic spectral clustering over confusion matrix for patch classification of Paris data set with all simple features. Edges width and colour are proportional to confusion. Manually drawn clusters for easier understanding.}{dim.fig:class-clustering-all-features}{0.95}
		
		Choosing which level of the class hierarchy to use depends on data set and applications.
		In a canonical classification perspective, we have to strongly reduce the number of classes if we want to have significant results.
		However reducing the number of class (i.e use a higher level in the classes hierarchy) also means that classes are more heterogeneous.
		  
		Both data set are extremely unbalanced (factor 100 or more). Thus our simple and direct Random Forest approach is ill suited for dealing with extremely small classes. (Cascading or one versus all framework would be needed).
		
		For Vosges data set a short analysis convince us to use 3 classes: Forest, Land, and other, on this three classes, the Land class is statistically overweighted by the 2 others.
		
		For the Paris data set, we again use a spectral layout to represent the affinity graph (See Figure \ref{dim.fig:class-clustering-all-features}).
		Introducing other features clearly helps to separate more classes.
		The graph also shows the limit of the classification, because some class cannot be properly defined without context (e.g. the side-walk, which is by definition the space between building and road, hence is defined contextually). 
		
		\subsubsection{Classification results}
		\myimage{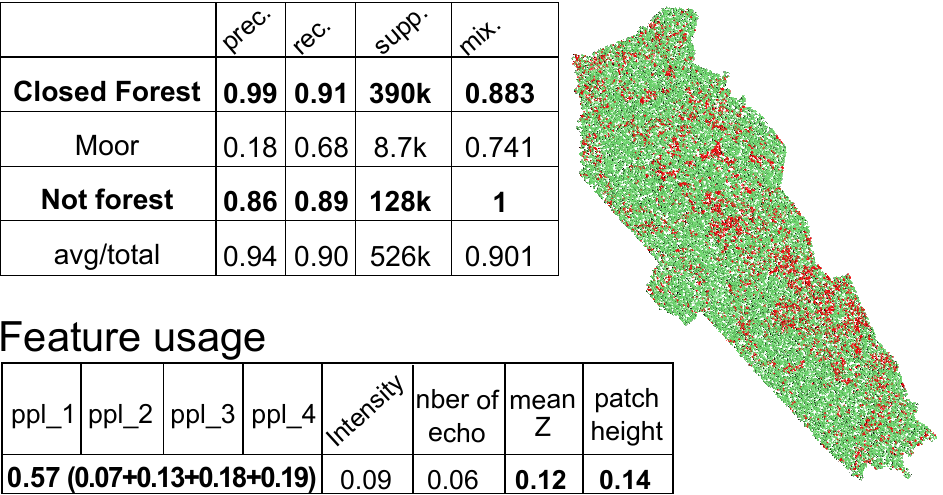}{Vosges dataset. (table 2) Precision(prec.), recall (rec.), support (supp.), and average percent of points of the class in the patches, for comParison with point based method (mix.). (table 1)Feature usage }{dim.fig:result-Vosges}{0.75}

		We perform a analysis of error on Vosges dataset and we note that errors seem to be significantly correlated to distance to borders. 
				 
		\myimageFullPageWidth{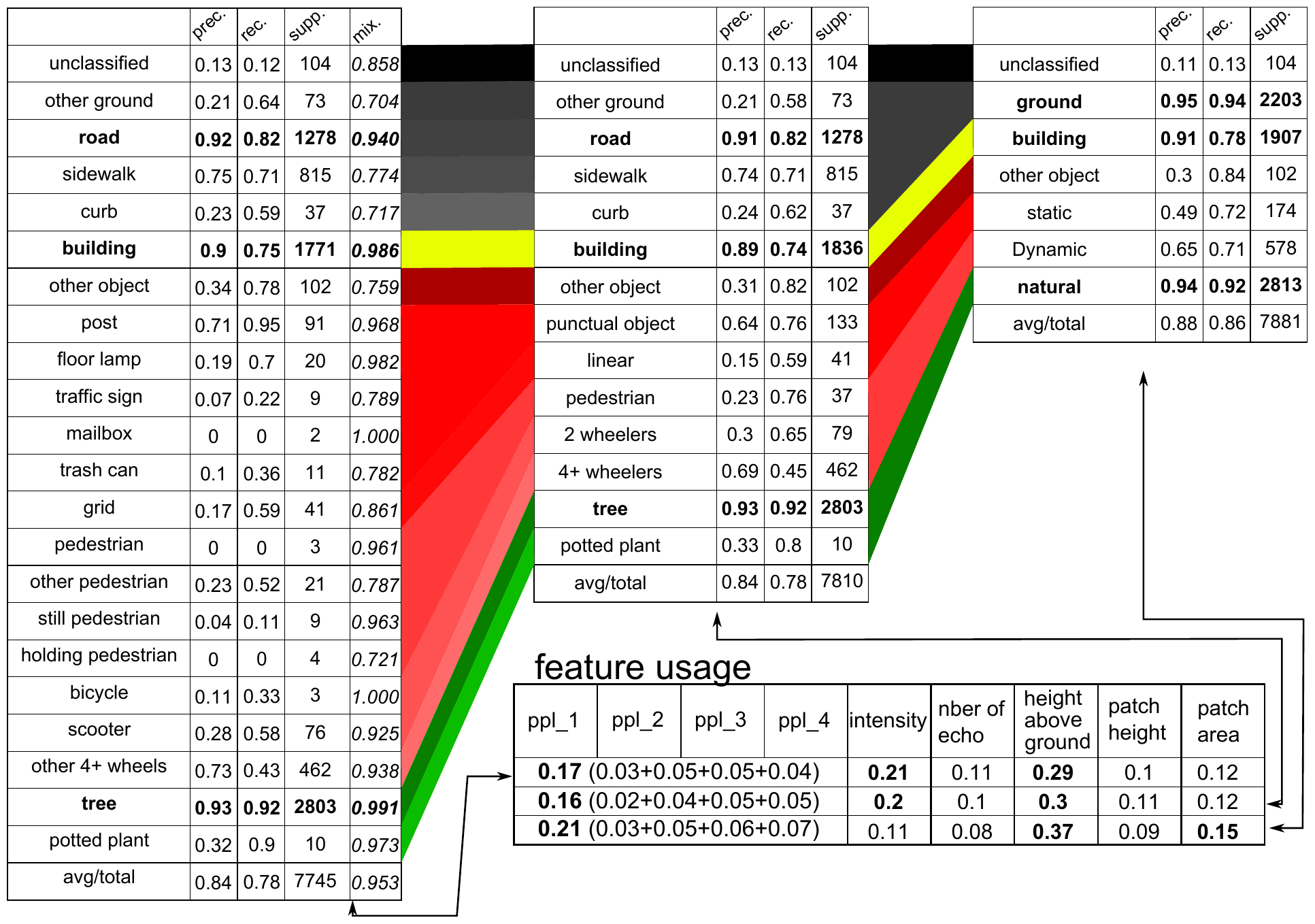}{Results for Paris data set: at various level of class hierarchy. Precision(prec.), recall (rec.), support (sup.) and average percent of points of the class in the patches of the class, for comParison with point based method (mix.). Classes of the same type are in the same continuous tone. Feature usage is evaluated for each level in the class hierarchy.}{dim.fig:result-Paris}{0.85}
		The learning time is less than a minute, the predicting time is less than a second. 
		
		For both dataset, patches main contain points from several classes. We measure how much of the patch points pertain to the dominant class. The result is given in the columns "mix". For instance the patch of the class "building" contains an average of 98.6 \% points of the class "building", whereas the patch from the class "forest" contains 88.3\% points of the class "forest".
		Therefore, to provide a comParison with point based classification, we can compute the precision of the classification per point as $Precision_{point} = Precision_{patch} * Mix.$. (Same for recall).
		
		\subsubsection{Precision or Recall increase} 
		\myimage{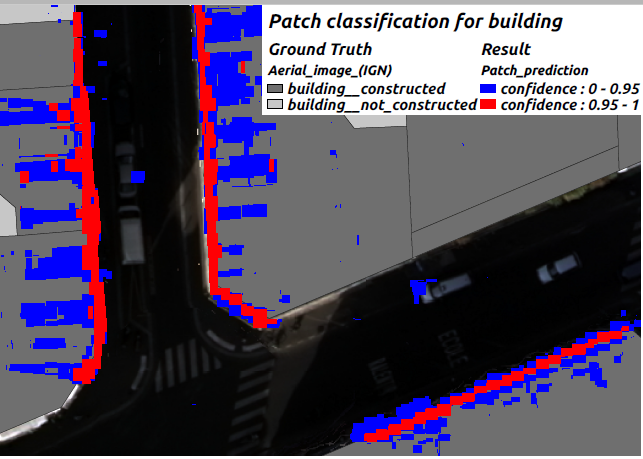}{Plotting of patches classified as building, using confidence to increase precision. Ground truth from IGN and Open Data Paris}{dim.fig:precision-increase}{0.75}
		
		As explained in Section \ref{dim.method.classification.using_confidence}, we can leverage the random forest confidence score to artificially increase the precision.
	
		We focus on the building class.
		As seen in the Figure \ref{dim.fig:precision-increase}, initial classification results (blue) are mostly correct.
		Yet, only keeping patches with high confidence may greatly increase precision (to $100$\%).
		Further filtering on confidence can not increase precision, but will reduce the variability of the found building patches. 
		This result (red) would provides a much better base for building reconstruction for instance. 
				
		\myimage{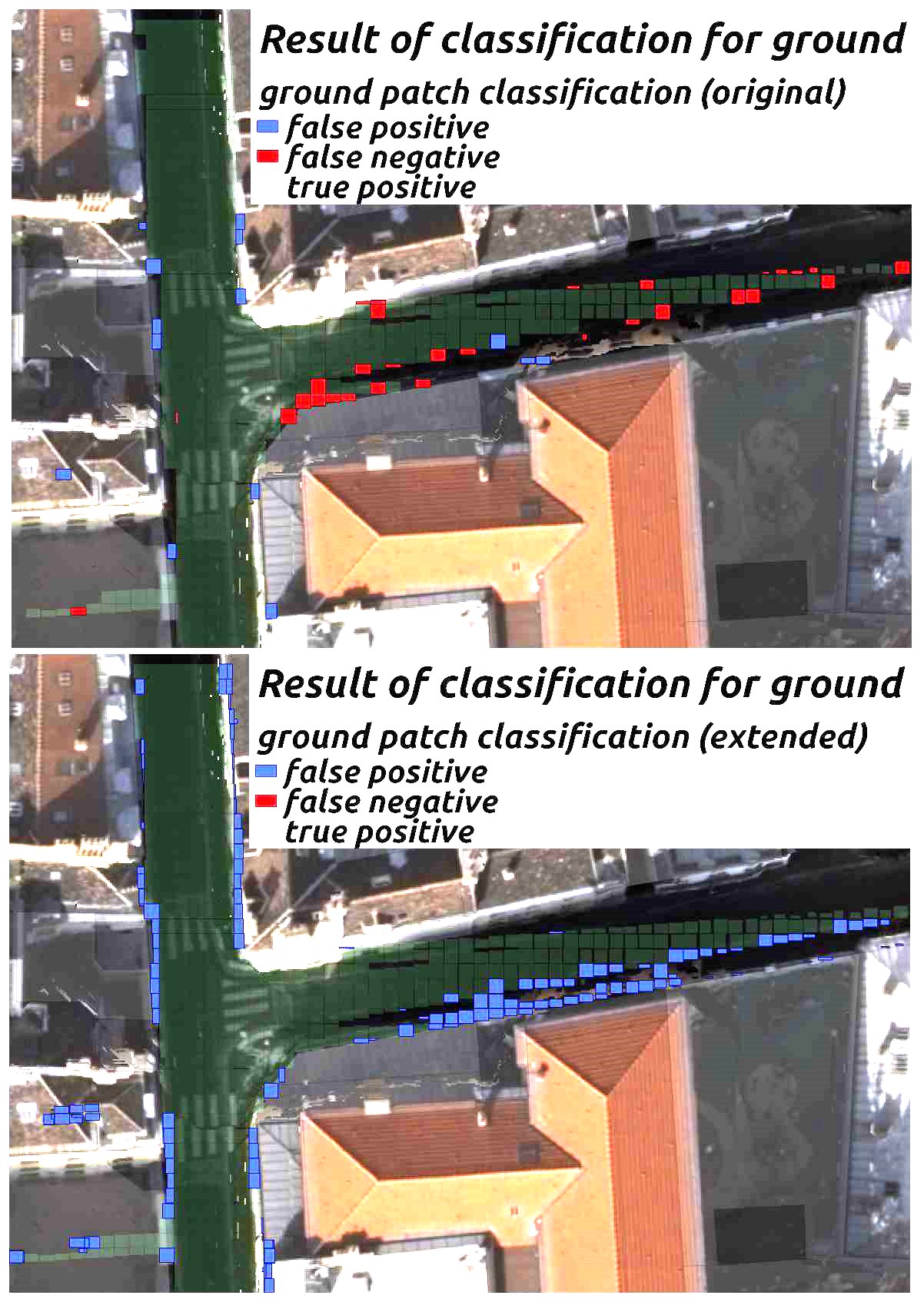}{Map of patch clustering result for ground. The classical result finds few extra patches that are not ground (blue), and misses some ground patches (red). Recall is increased by adding to the ground class all patches that are less than 2 meters in X,Y and 0.5 meter in Z around the found patches. Extra patches are much more numerous, but all the ground patches are found.}{dim.fig:recall-increase}{0.75}

		The patch classifier can also be used as a filtering preprocess.
		In this case, the goal is not to have a great precision, but to be fast and with a good recall.
		Such recall may be increased artificially for class of objects bigger than the sampling size ($1\cubic\metre$ for Paris).
		
		We take the example of ground classification (See Figure \ref{dim.fig:recall-increase}). 
		The goal is to find all ground patches very fast.
		We focus on a small area for illustration purpose. This area contains $3086$ patches, including $439$ ground patches.
		Patch classification finds $421$ ground patch, 
		with a recall of $92.7$\%.
		Using the found patch, all the surrounding patches (X,Y : $2$ \meter, Z : $0.5$ \meter ) are added to the result (few seconds).
		There are now $652$ patches in the result, and the recall is $100$\%.
		This means that from a filtering point of view, a complex classifier that would try to find ground points can be used on $652/3086=21\%$ of the data set, at the price of few seconds of computing, without any loss of information.



 \section{ Discussion }
	 \label{dim.sec:discussion} 
	 
	 \subsection{Point cloud server}
	\label{dim.par:pointcloudserver-limitation}
	We refer the reader to \cite{Cura2015} for an exhaustive analyse of the Point Cloud Server.
	Briefly, the PCS has demonstrated all the required capacities to manage point clouds and scale well.
	To the best of our knowledge the fastest and easiest way to filter very big point cloud using complex spatial and temporal criteria, as well as natively integrate point cloud with other GIS data (raster, vector).
	The main limitation is based on the hypothesis than points can be regrouped into meaningful (regarding further point utilisation) patches. If this hypothesis is false, the PCS lose most of its interest.
	 
	\subsection{Multi-scale local dimensionality descriptor}	 
		\myimageHL{"./illustrations/chap2/comparing_dim_desc/hist_comparison_for_tree"}{Histogram of $Dim_{LOD}$ and $Dim_{cov}$ for patch in trees (500 \kilo pts). Tree dimension could be from 1.2 to 2.6, yet $Dim_{LOD}$ is less ambiguous than $Dim_{cov}$}{dim.fig:hist_comparison_for_tree}{0.95}
		
		Tree patches are challenging for both dimensionality descriptor.
		There possible dimension changes a lot (See Fig. \ref{dim.fig:hist_comparison_for_tree}), although $Dim_{LOD}$ is more concentrated.
		Yet, $ppl$ is extremely useful to classify trees.
		Indeed, $ppl$ contains the dimensionality at various scale, and potentially the variation of it, which is quite specific for trees (See Fig. \ref{dim.fig:analysing_tree}).

		We stress that a true octree cell occupancy (i.e. without picking points as in the $ppl$) can be obtained without computing the octree, simply by using the binary coordinates.
		We implement it in python as a proof of concept. Computing it is about as fast as computing $Dim_{cov}$.
		
		Overall, $ppl$ offers a good alternative to the classical dimensionality descriptor ($Dim_{cov}$), being more robust and multi-scale. 
		However the $ppl$ also has limitations. 
		First the quality of the dimensionality description may be affected by a low number of points in the patch. 
		Second in some case it is hard to reduce it to a meaningful $Dim_{LOD}]$.
		Last because of assumption on density, it is sensible to geometric noise.
			  
	\subsection{Patch Classification}
		The $ppl$ descriptor contains lots of information about the patch. This information is leveraged by the Random Forest method and permit a rough classification based on geometric differences.
		As expected, $ppl$ descriptor are  not sufficient to correctly separate complex objects,
		which is the main limitation for a classification application. 
		
		The additional features are extremely simple, and far from the one used in state of the art.
		Notably, we don't use any contextual feature.
		We choose to classify directly in N classes, whereas due to the large unbalance in dataset, cascade or 1 versus all approaches would be more adapted.
		
	\subsubsection{Analysing class hierarchy} 
		The figure \ref{dim.fig:class-clustering-all-features} shows the limit of a classification without contextual information. For instance the class grid and buildings are similar because in Paris buildings balcony are typically made of grids.
		
		To better identify confusion between classes, we use a spectral layout on the affinity matrix.
		Graphing this matrix in 2D amount to a problem of dimensionality reduction.
		It could use more advanced method than simply using the first two eigen vector,
		in particular the two vector wouldn't need to be orthogonal (for instance, like in \cite{Hyvarinen2000}).
				 
	\subsubsection{Classification results}
		  First the feature usage for Vosges data set clearly shows that amongst all the simple descriptor, the $ppl$ descriptor is largely favoured.
		  This may be explained by the fact that forest and bare land have very different dimensionality, which is conveyed by the $ppl$ descriptor.
		  
		  Second the patch classifier appears to have very good result to predict if a patch is forest or not. The precision is almost perfect for forest. We reach the limit of precision of ground truth.
		  Because most of the errors are on border area, the recall for forest can also be easily artificially increased. The percent of points in patch that are in the patch class allow to compare result with a point based method. 
		  For instance the average precision per point for closed forest would be $0.99*0.883=0.874$ . We stress that this is averaged results, and better precision per point could be achieved because we may use random forest confidence to guess border area (with a separate learning for instance).
		  For comParison with point methods, the patch classifier predict with very good precision and support over 6 billions points in few seconds (few minutes for training). We don't know other method that have similar result while being as fast and natively 3D.
		   The Moor class can't be separated without more specialised descriptor, because Moor and no forest classes are geometrically very similar.
		  
		  The principal limitation is that for this kind of aerial Lidar data set the 2.5D approximation may be sufficient, which enables many raster based methods that may perform better or faster.
		  
		  The figure \ref{dim.fig:result-Paris} gives full results for Paris data set, at various class hierarchy level.
		  Because the goal is filtering and not pure classification, we only comment the 7 classes result. The proposed methods appears very effective to find building, ground and trees.
		  Even taking into consideration the fact that patch may contains mixed classes (column mix.), the result are in the range of state of the art point classifier, while being extremely fast. 
		  This result are sufficient to increase recall or precision to 1 if necessary.
		  We stress that even results appearing less good (4+wheelers , 0.69 precision, 0.45 recall) are in fact sufficient to increase recall to 1 (by spatial dilatation of the result), which enables then to use more subtle methods on filtered patches.
		  
		  $ppl$ descriptor is less used than for the Vosges data set, but is still useful, particularly when there are few classes.
		  It is interesting to note that the mean intensity descriptor seems to be used to distinguish between objects, which makes it less useful in the 7 classes case.
		  The patch classifier for Paris data set is clearly limited to separate simple classes. In particular, the performances for objects are clearly lower than the state of the art. A dedicated approach should be used (cascaded or one versus all classifier).

	  \subsubsection{Estimating the speed and performance of patch based classification compared to point based classification}
		  The Point Cloud Server is designed to work on patches, which in turns enable massive scaling. 
		  
		  Timing a server is difficult because of different layer of caches, and background workers. Therefore, timing should be considered as order of magnitude.
		  For Paris data set,extracting extra classification features requires $\sim \frac{400 \second}{n_{workers}}$( 1 to 8 workers), learning $\sim 210 \second$,
		  and classification $\sim$ few \second.
		  We refer to \cite{Weinmann2015}(Table 5) for point classification timing on the same dataset (4.28\hour, 2\second, 90\second ) (please note that the training set is much reduced by data balancing).
		  As expected the speed gain is high for complex feature computing (not required) and point classification (done on patch and not points in our case).
		  
		  For Vosges data set, features are extracted at $1 \mega pts \per \second \per worker$, learning $\sim few \minute$, classification $\sim 10 \second$.
		  The Vosges data set has not been used in other articles, therefore we propose to compare the timings to \cite{shapovalov2010} (Table 3). Keeping only feature computation and random forest training (again on a reduced data set), they process 2 \mega points in 2 \minute, whereas our method process the equivalent of 5.5 B points in few minutes.

		  Learning and classification are mono-threaded (3 folds validation), although the latter is easy to parallelise.
		  Overall, the proposed method is one to three orders of magnitude faster.
		  
		  For Paris data set (Fig. \ref{dim.fig:result-Paris}), we compare to \cite{Weinmann2015}(Table 5). As expected there results are better, particularly in terms of precision (except for the class of vegetation). This trend is aggravated by taking into account the "mix." factor.
		  Indeed we perform patch classification, and patch may not pertain to only one class, which is measured by the mix factor (amount of points in the main class divided by the total number of point).
		  However, including the mix factor the results are still within the 85 to 90 \% precision for the main classes (ground, building, natural).
		  
		  For Vosges data set (Fig \ref{dim.fig:result-Vosges}), we refer to \cite{shapovalov2010} (Table 2). There random forest classifier get trees with 93\% precision and  89\% recall.
		  Including the mix factor we get trees with a precision of 87\% and 80\% recall.
		  As a comParison to image based classification, an informal experiment of classification with satellite image reaches between 85 \% and 93 \% of precision for forest class depending on the pixel size (between 5 and 0.5 \metre).
		  
		  Overall, the proposed method get reasonably good results compared to more sophisticated methods,
		  while being much faster.
		  It so makes a good candidate as a preprocessing filtering step.
		  
	 \subsubsection{Precision or Recall increase}
	 Because the propose methods are pre-process of filtering step, it can be advantageous to increase precision or recall.
		
		\myimageHL{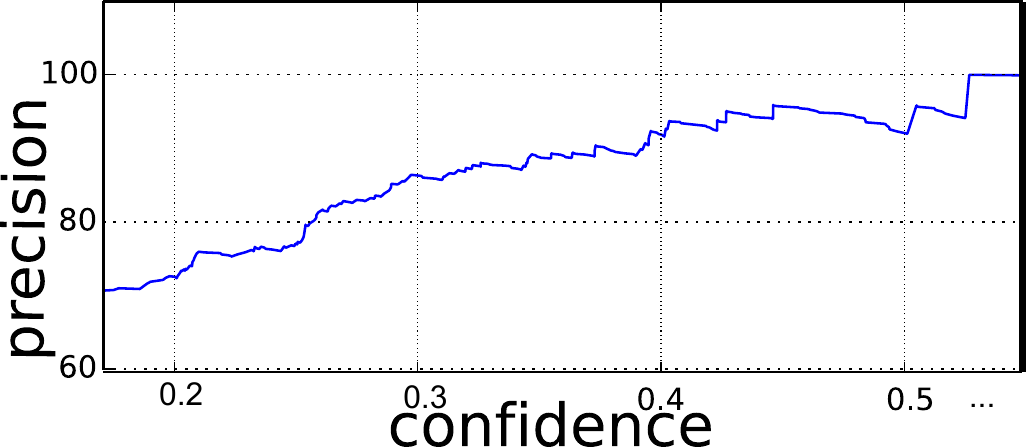}{Precision of 4+wheelers class is a roughly rising function of random forest confidence scores.}{dim.fig:precision-vs-confidence}{0.95}
		In the Figure \ref{dim.fig:precision-increase} gives a visual example where increasing precision and reducing class heterogeneity is advantageous. This illustrates that having a $1$ precision or recall is not necessary the ultimate goal.
		In this case it would be much easier to perform line detection starting from red patches rather than blue patches.
		\\
		The limitation of artificial precision increase is 
		that it is only possible when precision is roughly a rising function of random forest confidence, as seen on the illustration \ref{dim.fig:precision-vs-confidence}.
		For this class, by accepting only prediction of random forest that have a confidence over $0.3$ the precision goes from $0.68$ to $0.86$, at the cost of ignoring half the predictions for this class.
		This method necessitates that the precision is roughly a rising function of the confidence, as for the 4+wheeler class for instance (See Figure \ref{dim.fig:precision-vs-confidence}).   
		This strategy is not possible for incoherent class, like unclassified class.

		The method we present for artificial recall increase is only possible if at least one patch of each object is retrieved, and objects are spatially dense.
		This is because a spatial dilatation operation is used.
		This is the case for "4+wheelers" objects in the Paris data set for instance.
		The whole method is possible because spatial dilatation is very fast in point cloud server (because of index).
		Moreover, because the global goal is to find all patches of a class while leaving out some patches,
		it would be senseless to dilate with a very big distance.
		In such case recall would be $1$, but all patches would be in the result, thus there would be no filtering, and no speeding.
		\\
		The limitation is that this recall increase method is more like a deformation of already correct results 
		rather than a magical method that will work with all classes.
				


\section{Conclusion} 
	We propose a dimensionality descriptor based on octree cells occupancy at different scales. It allows to describe local geometrical dimensionality of a point cloud in a meaningful way, and concentrates information, which can be leveraged for helping visualisation, or by a rough classifier.
	This descriptor is less affected by sampling than alternative structure tensor geometric descriptor, fast to compute (a simple ordering of the points), and intrinsically multi-scales.
	We show the interest of this descriptor, both by comParison to the state of the art dimensionality descriptor, and by proof of its usefulness in real Lidar dataset classification.
	Classification is extremely fast, sometime at the price of performance (precision / recall). 
	However we prove that those results can be used as a pre-processing step for more complex methods, using if necessary precision-increase or recall-increase strategies.



	\section{Bibliography}  
	\bibliography{./implicit_LOD} 

%
%
%
%

\end{document}